\documentclass{article}
\usepackage{ijcai26}
\usepackage{times}
\usepackage{soul}
\usepackage{url}
\usepackage[hidelinks]{hyperref}
\usepackage[utf8]{inputenc}
\usepackage[small]{caption}
\usepackage{graphicx}
\usepackage{amsmath}
\usepackage{amsthm}
\usepackage{booktabs}
\usepackage{algorithm}
\usepackage{algorithmic}
\usepackage[switch]{lineno}
\usepackage{amsfonts}

\author{
Jingtao Wang$^{1,2\dagger}$
Yucong Wang$^{3,4\dagger}$
Jun Ding$^{1,2,5,6,7}$
Rui Cai $^{3,4,*}$\And
Xun Wang $^{3,4,*}$
\affiliations
$^1$Meakins-Christe Laboratories, Research Institute of McGill University Health Centre, 1001 Decarie Blvd, Montreal, H4A 3J1, Quebec, Canada\\
$^2$Department of Medicine, Division of Experimental Medicine, McGill University, 1001 Decarie Blvd, Montreal, H4A 3J1, Quebec, Canada\\
$^3$The College of Computer Science and Technology, Zhejiang Gongshang University, Hangzhou, China\\
$^4$Zhejiang Key Laboratory of Big Data and Future E-Commerce Technology, Hangzhou, China\\
$^5$Quantitative Life Sciences, McGill University, 845 Rue Sherbrooke Ouest, Montreal, H3A 0G4, Quebec, Canada\\
$^6$School of Computer Science, McGill University, 3480 Rue University, Montreal, H3A 2A7, Quebec, Canada\\
$^7$Mila-Quebec AI Institute, 6666 Rue Saint-Urbain, Montreal, H2S 3H1, Quebec, Canada\\
\thanks{$\dagger$ These authors contributed equally.}
\thanks{$*$ Corresponding authors.}
\emails
cairuics@gmail.com,
wx@zjgsu.edu.cn
}

\title{Summarize Before You Speak with ARACH: A Training-Free Inference-Time Plug-In for Enhancing LLMs via Global Attention Reallocation}

\begin{document}

\maketitle

\begin{abstract}
    Large language models (LLMs) achieve remarkable performance, yet further gains often require costly training. This has motivated growing interest in post-training techniques—especially training-free approaches that improve models at inference time without updating weights. Most training-free methods treat the model as a black box and improve outputs via input/output-level interventions, such as prompt design and test-time scaling through repeated sampling, reranking/verification, or search. In contrast, they rarely offer a plug-and-play mechanism to intervene in a model’s internal computation. We propose 
    \textbf{ARACH} (\textbf{A}ttention \textbf{R}eallocation via an \textbf{A}daptive \textbf{C}ontext \textbf{H}ub),
    a training-free inference-time plug-in that augments  
    LLMs with an adaptive context hub to aggregate context and reallocate attention. Extensive experiments across multiple language modeling tasks show consistent improvements with modest inference overhead and no parameter updates. Attention analyses further suggest that ARACH mitigates the attention sink phenomenon. These results indicate that engineering a model's internal computation offers a distinct inference-time strategy, fundamentally different from both prompt-based test-time methods and training-based post-training approaches.

\end{abstract}

\section{Introduction}

Large language models (LLMs) have demonstrated strong performance across a wide range of tasks, from open-ended generation to question answering and long-form modeling.~\cite{mann2020language} This progress is largely driven by scaling model size and data, as well as post-training alignment and adaptation~\cite{kaplan2020scaling,hoffmann2022training,rafailov2023direct}. However, further improving a deployed LLM is often expensive: continued pretraining, task-specific finetuning, and alignment pipelines can require substantial compute, engineering effort~\cite{han2024parameter}, and in some settings access to proprietary data or human feedback~\cite{DBLP:journals/corr/abs-2203-02155}.


A major direction for improving LLMs is \emph{post-training optimization} that updates parameters. This includes supervised instruction tuning and alignment with human preferences, typically implemented via RLHF-style pipelines, and a broad family of parameter-efficient finetuning (PEFT) methods such as adapters~\cite{houlsby2019parameter}, prefix/prompt tuning~\cite{li-liang-2021-prefix,lester2021prompttuning}, and low-rank adaptation~\cite{hu2022lora}. These approaches are powerful, but they still require training-time resources, careful hyperparameter selection, and versioned deployment of new weights~\cite{dettmers2023qlora,zhang2023adalora}. 

A second direction is \emph{training-free inference-time improvement} via prompting and test-time scaling.
Few-shot prompting~\cite{mann2020language} and chain-of-thought prompting~\cite{wei2022chain} can improve reasoning and task performance without changing weights, 
while techniques such as self-consistency~\cite{wang2022self}, multi-sample decoding with reranking/verification~\cite{cobbe2021training}, and tree-of-thought~\cite{yao2023tree} style search allocate additional compute at inference to increase reliability. 
Despite their effectiveness, most of these methods act purely in the \emph{input/output space}: they treat the model as a black box and rely on repeated sampling, longer prompts, or external scoring~\cite{zhang2025survey}. 
Consequently, they may incur significant test-time overhead and rarely provide a plug-and-play mechanism to intervene in the model's \emph{internal computation}.

To bridge this gap, we explore a complementary, training-free direction: intervening in the model's \emph{internal computation} at inference time.
Our method, 
ARACH (Attention Reallocation via an Adaptive Context Hub),
is a inference time plug-in for decoder-only Transformers.
At its core, ARACH augments LLM with a lightweight, adaptive \textit{context hub}, which could dynamically summarize the causally available prefix and provide an additional conditioning path for next-token prediction. 
Specifically, this hub operates as a parallel stream of tokens that runs alongside the standard verbal token stream. 
Under strict causal constraints, each position in this hub stream dynamically aggregates and summarizes information from the entire causally available prefix up to that point. 
This design provides the model with an explicit, compact, and readily accessible representation of long-range context during next-token prediction. 
By introducing this additional computational pathway within the attention mechanism itself, ARACH enables a form of internal \textit{summarize-then-generate} reasoning without modifying any pre-trained model weights, offering a truly plug-and-play enhancement that can be toggled on or off at inference time. 

To ensure balanced and effective attention reallocation, ARACH incorporates a  tunable logit offset applied to hub-related connections. 
Without this regulation, the new hub pathway could attract excessive attention mass, leading to a \textit{routing collapse} where the original context is underutilized—an issue analogous to the attention sink phenomenon~\cite{xiao2024efficient}. 
The single scalar offset term serves as a calibration knob, directly modulating the influence of hub within the softmax operation. 
This stabilizes inference by ensuring a productive division of attention between the novel hub-mediated route and the standard token-to-token interactions.
Ultimately, the synergy between the adaptive context hub and its logit offset forms the core of ARACH, enabling systematic improvements in next-token prediction through engineered, internal attention redistribution.



We evaluate ARACH under paired settings where weights and decoding configurations are fixed and only ARACH is toggled on/off.
Across language modeling and cloze-style benchmarks, ARACH yields consistent gains with limited additional computation.
Attention analyses further indicate reduced sink-like attention concentration and increased reliance on hub-mediated routing.

Our main contributions are summarized as follows:
\begin{itemize}

  \item  We introduce ARACH,  a training-free, inference-time plug-in that equips decoder-only Transformers with an adaptive context hub. It operates under strict causal constraints and uses a lightweight logit offset to steer the hub’s attention routing.
  \item  In paired evaluations with fixed model backbones and decoding setups, ARACH achieves consistent gains across multiple language modeling and cloze-style benchmarks. 
  \item Our attention-based analysis reveals that ARACH enhances hub-mediated context aggregation and alleviates sink-like attention concentration, providing mechanistic insight into its performance gains. 
  
\end{itemize}

\section{Related Work}

\subsection{Training-Based LLMs Adaptation}
A common way to improve LLMs is to update parameters after pretraining.
Full finetuning and supervised instruction tuning can substantially boost task-following behavior, and alignment pipelines based on human feedback (e.g., RLHF~\cite{DBLP:journals/corr/abs-2203-02155}) further optimize responses toward human preferences~\cite{chung2024scaling}.
While effective, these methods require training compute and careful engineering, and they produce new model versions that must be stored, served, and audited~\cite{han2024parameter}.

Parameter-efficient finetuning (PEFT) reduces the cost of adaptation by updating only a small set of parameters while freezing most pretrained weights~\cite{han2024parameter}.
Representative approaches include adapters~\cite{houlsby2019parameter}, prefix tuning~\cite{li-liang-2021-prefix}, prompt tuning~\cite{lester-etal-2021-power}, and low-rank adaptation (LoRA)~\cite{hu2022lora} variants. These methods can be competitive with full finetuning, but they still require training, hyperparameter selection, and maintaining task-specific artifacts~\cite{dettmers2023qlora}.

Despite their success, all the approaches above rely on parameter updates and thus incur non-trivial training overhead.
This motivates a complementary line of work that seeks to improve model behavior without modifying parameters, by intervening directly in inference-time computation.

\subsection{Training-free Inference-time Methods}
Another line of work improves performance without changing weights, by intervening in the input/output process.
Few-shot prompting~\cite{mann2020language} and prompt engineering exploit in-context learning, while chain-of-thought~\cite{wei2022chain} prompting encourages intermediate reasoning traces.
Test-time scaling methods further allocate compute during inference using repeated sampling and selection.
For example, self-consistency~\cite{wang2022self} aggregates multiple reasoning paths to improve robustness, while reranking and verification ~\cite{DBLP:journals/corr/abs-2203-02155,cobbe2021training} leverage auxiliary signals (e.g., reward models or self-critique) to select better candidates. 
Furthermore, tree-of-thought~\cite{yao2023tree} style methods perform explicit search over intermediate reasoning states.

These approaches are attractive because they are training-free and often model-agnostic~\cite{zhang2025survey}.
However, they typically treat the model as a black box and operate mainly in the input space (longer prompts) or output space (sampling, reranking, verification), which may incur substantial inference overhead~\cite{zhang2025survey}.
In contrast, ARACH provides a plug-and-play mechanism to modify \emph{internal attention routing} deterministically, offering an orthogonal axis that can potentially be combined with prompt-based test-time scaling.

\subsection{Attention Sink Phenomenon}
Recent analyses~\cite{xiao2024efficient} reveal a sink-like phenomenon in long-context modeling, where early tokens attract disproportionately large attention mass despite providing limited semantic information.
This effect has been leveraged in streaming deployment settings~\cite{xiao2024efficient}, but also highlights a broader issue: attention can be misallocated in ways that reduce effective context utilization~\cite{liu2024lost}.
These findings motivate studying inference-time mechanisms that reallocate attention toward more useful context summaries, which is closely related to the hub-mediated routing design explored in our work.

\begin{figure*}[!htb]
  \centering
  \setlength{\belowcaptionskip}{0pt}
  \includegraphics[width=\textwidth,trim=4mm 205mm 4mm 0mm,]{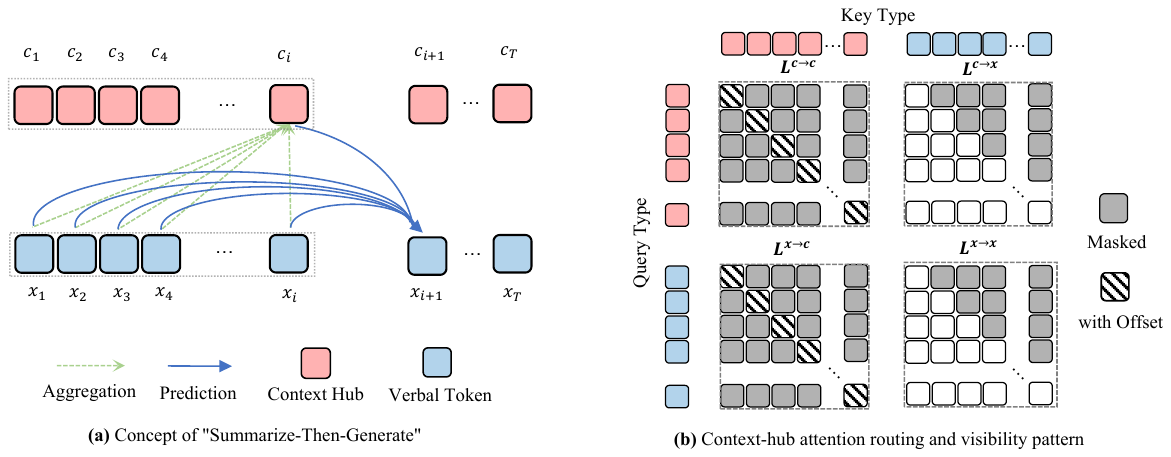} 
    \caption{
    \textbf{ARACH overview.}
    \textbf{(a)} \textit{Summarize-then-generate} intuition.
    At each decoding step, an index-aligned \emph{context hub stream} $\{c_i\}$ aggregates the causally available prefix and supports next-token prediction alongside the verbal token stream $\{x_i\}$.
    \textbf{(b)} Hub-based attention routing realized by a two-stream attention layout.
    Self-attention is partitioned into 
    four blocks under hub-specific visibility constraints.
    A scalar logit offset $b$  is applied to regulate the strength of hub-mediated attention reallocation at inference time.
    }
      \label{fig:method_overview}
\end{figure*}

\section{Method - ARACH}
\subsection{Overview of ARACH}
ARACH is a training-free, inference-time plug-in for
\emph{pretrained decoder-only Transformer language models}.
It targets the common deployment setting where a model has already been trained, and one seeks to improve inference behavior \emph{without updating parameters} by intervening in the model's internal computation.

At a high level, ARACH augments standard autoregressive generation with a lightweight \emph{context hub}---implemented as a hub \emph{stream} that repeats a single \emph{frozen} hub token type across time steps---to reallocate attention toward a compact summary of the causally available prefix (Figure~\ref{fig:method_overview})
.
The hub provides an explicit aggregation route for long-range context, while remaining fully compatible with the original attention mechanism and autoregressive decoding process.
Importantly, ARACH operates solely at inference time and can be toggled on or off as a modification to the attention graph, without introducing additional training or parameter updates.

Concretely, ARACH consists of two modular components:
(i) a \emph{context hub} realized by a two-stream layout (a hub stream repeating a single frozen hub token type) with hub-specific visibility constraints, and
(ii) a lightweight \emph{hub-attention logit offset} that regulates the strength of hub-mediated attention reallocation.
We first describe the hub layout and its visibility constraints, and then introduce the logit-offset-based regulation mechanism.

\subsection{Notation and Preliminaries}
\label{sec:prelim}

\paragraph{Autoregressive setting.}
Let $x_{1:T}$ denote the sequence of \emph{verbal} tokens in a context window, where $T$ is the number of verbal tokens.
At step $i$, a decoder-only Transformer language model predicts the next token $x_{i+1}$ conditioned on the past (causally visible) context $x_{1:i}$.

\paragraph{Self-attention notation.}
For a transformer layer $\ell$ with $H$ heads, let
$Q^\ell, K^\ell, V^\ell \in \mathbb{R}^{B \times H \times T \times d_h}$
be the queries, keys, and values (with per-head dimension $d_h$), where $B$ is the batch size and $T$ denotes the length of the verbal sequence.
Baseline causal self-attention computes
\begin{equation}
\label{eq:selfatt}
\mathrm{Attn}(Q^\ell,K^\ell,V^\ell)
=
\mathrm{softmax}\!\left(\frac{Q^\ell (K^\ell)^\top}{\sqrt{d_h}} + M\right) V^\ell ,
\end{equation}
where $M \in \mathbb{R}^{T\times T}$ is the causal mask ($M_{ij}=0$ if $j\le i$ and $-\infty$ otherwise). 
The self-attention output is then obtained by the weighted sum of values:
\begin{equation}
O^\ell = A^\ell V^\ell = A^\ell\!\left(X^{\ell-1}W_V^\ell\right),
\end{equation}
where $X^{\ell-1}\in\mathbb{R}^{B\times T\times d}$ denotes the layer input and $W_V^\ell$ is the value projection.

\paragraph{Token streams.}
We distinguish between two types of tokens: verbal tokens, denoted by $x$, corresponding to the original language model inputs, and context hub tokens, denoted by $c$, introduced by ARACH. 
Conceptually, ARACH maintains a \emph{per-step} hub state that evolves during next-token prediction; in the prefill view, we represent these states as an index-aligned hub token stream $c_{1:T}$. 
We use superscripts $X$ and $C$ to indicate quantities associated with the verbal and hub streams, respectively.

\subsection{Global Aggregation with a Context Hub}
\label{sec:layout}

\noindent\textbf{Motivation and setting.}
Given a fixed pretrained decoder-only Transformer language model, ARACH introduces a \emph{context hub} to aggregate information from the causally visible past and make it readily accessible for next-token prediction (Figure~\ref{fig:method_overview}a).
Importantly, ARACH does \emph{not} modify model parameters; it only changes how self-attention is computed at inference time by augmenting the token layout and the attention mask.

\paragraph{Two-stream token layout.}
For an input context $x_{1:T}$, ARACH constructs an additional \emph{hub stream} $c_{1:T}$ with the same length and aligns hub token $c_i$ with verbal token $x_i$ at index $i$.
For clarity, we describe ARACH using a full-context (prefill) view in which the two streams are concatenated:
\begin{equation}
s = [\,c_{1:T}\,;\,x_{1:T}\,], \qquad |s|=2T.
\end{equation}

All hub positions share the same hub \emph{token type}, i.e., they use a single shared embedding across all time indices.
This embedding is initialized once at inference time by sampling from a Gaussian distribution whose mean and covariance match those of the pretrained token embedding matrix (mean-resizing initialization), and is then kept fixed throughout inference.
This ensures that the entire ARACH mechanism remains entirely training-free, with no model parameters updated or fine-tuned.

As a result, all hub tokens start from an identical representation; their step-wise semantic roles are induced entirely by the hub--verbal attention routing and causal visibility constraints described in Section~\ref{sec:routing}, rather than by position- or index-specific embeddings.

\paragraph{Position encoding.}
ARACH assigns a constant position index to all hub tokens while keeping standard monotonic positions for verbal tokens:
\begin{equation*}
\pi(c_i)=0,\qquad \pi(x_i)=i \ \ (i=1,\ldots,T).
\end{equation*}
Thus, hub tokens do not encode absolute positions; instead, their step-wise association is induced by the hub--verbal visibility pattern (in particular, the index-aligned diagonal connections) described in Section~\ref{sec:routing}.

\paragraph{What the hub represents.}
Under the routing constraints, each hub token $c_i$ aggregates information from the causally visible prefix (that is, $x_{1:i}$) through hub-mediated self-attention, yielding a compact context summary that can be attended to when predicting $x_{i+1}$.
ARACH introduces no additional learned modules: the hub representation emerges implicitly from the modified attention pattern.

\subsection{Hub-Based Attention Layout and Visibility}
\label{sec:routing}

\noindent\textbf{Goal.}
This subsection specifies how the context hub is incorporated into attention computation.
We first present the two-stream attention in block-matrix form, and then define a causal visibility mask that implements the hub-based routing shown in Figure~\ref{fig:method_overview}b.

\paragraph{Two-stream attention in block form.}
For a Transformer layer $\ell$ with $H$ heads, we denote the per-head queries, keys, and values for the hub stream and verbal stream as
\begin{align*}
Q^{C,\ell},K^{C,\ell},V^{C,\ell} &\in \mathbb{R}^{B \times H \times T \times d_h},\\
Q^{X,\ell},K^{X,\ell},V^{X,\ell} &\in \mathbb{R}^{B \times H \times T \times d_h}.
\end{align*}

Let $Q^{\ell}=[Q^{C,\ell};Q^{X,\ell}]$ (and similarly for $K^\ell,V^\ell$) denote the concatenated tensors aligned with $s=[c_{1:T};x_{1:T}]$.
The pre-softmax attention logits can be written as a $2\times2$ block matrix:
\begin{equation}
\label{eq:block_logits}
\begin{aligned}
\frac{Q^\ell (K^\ell)^\top}{\sqrt{d_h}}
&=
\begin{bmatrix}
\frac{Q^{C,\ell}(K^{C,\ell})^\top}{\sqrt{d_h}} &
\frac{Q^{C,\ell}(K^{X,\ell})^\top}{\sqrt{d_h}} \\
\frac{Q^{X,\ell}(K^{C,\ell})^\top}{\sqrt{d_h}} &
\frac{Q^{X,\ell}(K^{X,\ell})^\top}{\sqrt{d_h}}
\end{bmatrix}\\
&\equiv
\begin{bmatrix}
L^{C\to C,\ell} & L^{C\to X,\ell}\\
L^{X\to C,\ell} & L^{X\to X,\ell}
\end{bmatrix},
\end{aligned}
\end{equation}
where the superscript $A\!\to\!B$ indicates attention \emph{from} query stream $A$ (rows) \emph{to} key stream $B$ (columns).
We obtain ARACH attention weights by adding the block mask $M_{\textsc{arach}}$ (and later the logit offset) to these logits before the row-wise softmax.
Compared to the baseline decoder-only attention ($X\!\to\!X$ with a causal mask), ARACH enables three additional hub-related pathways: $C\!\to\!C$, $X\!\to\!C$, and $C\!\to\!X$.
The final attention weights are obtained by applying a row-wise softmax over the \emph{permitted} keys under the ARACH mask.

\paragraph{Visibility pattern (four-quadrant mask).}
ARACH enforces the following visibility rules for each time index $i$:


\begin{itemize}\setlength{\itemsep}{0pt}
  \item \textbf{$C\!\to\!C$ (hub $\to$ hub):} diagonal-only ($c_i\!\to\!c_i$).
  \item \textbf{$X\!\to\!C$  (verbal $\to$ hub):} diagonal-only ($x_i\!\to\!c_i$).
  \item \textbf{$C\!\to\!X$ (hub $\to$ verbal):} causal ($c_i\!\to\!x_{1:i}$).
  \item \textbf{$X\!\to\!X$ (verbal $\to$ verbal):} standard causal ($x_i\!\to\!x_{1:i}$).
\end{itemize}

\noindent
These rules are expressed by a single block-structured mask
$M_{\textsc{arach}}\in\mathbb{R}^{2T\times2T}$:
\begin{equation}
\label{eq:arach_mask}
M_{\textsc{arach}}=
\begin{bmatrix}
M_{\mathrm{diag}} & M_{\mathrm{causal}}\\
M_{\mathrm{diag}} & M_{\mathrm{causal}}
\end{bmatrix},
\end{equation}
where $(M_{\mathrm{diag}})_{ij}=0$ if $j=i$ and $-\infty$ otherwise, and
$(M_{\mathrm{causal}})_{ij}=0$ if $j\le i$ and $-\infty$ otherwise.
This four-quadrant structure implements the hub-based routing in
Figure~\ref{fig:method_overview}b while preserving autoregressive causality.

\begin{table*}[t]
\centering
\setlength{\abovecaptionskip}{2pt}
\setlength{\belowcaptionskip}{0pt}
\small
\setlength{\tabcolsep}{4.2pt}
\begin{tabular}{lcccccccc}
\toprule
\textbf{Dataset} & \textbf{Metric} & \textbf{Baseline}
& \multicolumn{2}{c}{$b=-0.3$}
& \multicolumn{2}{c}{$b=-0.4$}
& \multicolumn{2}{c}{$b=-0.5$} \\
\cmidrule(lr){4-5}\cmidrule(lr){6-7}\cmidrule(lr){8-9}
& & & \textbf{+ARACH} & \textbf{$\Delta$}
    & \textbf{+ARACH} & \textbf{$\Delta$}
    & \textbf{+ARACH} & \textbf{$\Delta$} \\
\midrule
LAMBADA      & Acc (\%) $\uparrow$ & 46.89 & 49.93 & +3.04 & 50.24 & +3.35 & \textbf{50.42} & +3.53 \\
StoryCloze   & Acc (\%) $\uparrow$ & 57.46 & 57.51 & +0.05 & \textbf{57.56} & +0.10 & \textbf{57.56} & +0.10 \\
SQuAD        & EM $\uparrow$        & 50.41 & \textbf{50.90} & +0.49 & 50.73 & +0.32 & 50.54 & +0.13 \\
SQuAD        & F1 $\uparrow$        & 59.71 & \textbf{60.48} & +0.77 & 60.32 & +0.61 & 60.18 & +0.47 \\
WikiText-103 & PPL $\downarrow$     & 29.37 & \textbf{29.17} & +0.20 & 29.21 & +0.16 & 29.29 & +0.08 \\
PG-19        & PPL $\downarrow$     & 37.33 & 34.60 & +2.73 & 33.82 & +3.51 & \textbf{33.11} & +4.22 \\
\bottomrule
\end{tabular}
\caption{
Paired evaluation of GPT-2 small with and without ARACH under identical settings (no weight updates).
We report results for $b\in\{-0.3,-0.4,-0.5\}$.
We define $\Delta$ such that positive values indicate improvement.
}

\label{tab:main_results}
\end{table*}

\subsection{Calibrating Hub Attention with a Logit Offset}
\label{sec:attenreg}

\noindent\textbf{Why an offset term.}
The hub stream provides an additional route for aggregating and retrieving information from the causally visible prefix. However, the \emph{strength} of interactions that involve the hub should be tunable at inference time.
Without an explicit control, attention can over-concentrate on these shortcut connections, leading to a routing collapse in which the hub becomes an attractor and effective utilization of the original context is reduced (a sink-like concentration that can emerge in long-context decoding).
We therefore introduce a single scalar pre-softmax logit offset $b$ to calibrate hub-related attention strength without changing model parameters.

\paragraph{Hub-attention logit offset.}
The offset $b$ is added to the pre-softmax logits of the hub-diagonal connections
in the $C\!\to\!C$ and $X\!\to\!C$ blocks, i.e., to $(c_i\!\to\!c_i)$ and
$(x_i\!\to\!c_i)$ for all $i\in\{1,\ldots,T\}$.
We implement this by an additive matrix $B(b)\in\mathbb{R}^{2T\times2T}$:
\begin{equation}
\label{eq:bias_matrix}
B(b)=
\begin{bmatrix}
bI_T & 0\\
bI_T & 0
\end{bmatrix},
\end{equation}
where $I_T$ is the $T\times T$ identity matrix.
For each head $h$, ARACH attention is computed as
\begin{equation}
\begin{aligned}
\label{eq:arach_attn}
A^{\ell,h}
&=
\mathrm{softmax}\!\left(
\frac{Q^{\ell,h}(K^{\ell,h})^\top}{\sqrt{d_h}}
+ M_{\textsc{arach}}
+ B(b)
\right),\qquad \\
O^{\ell,h}&=A^{\ell,h}V^{\ell,h}.
\end{aligned}
\end{equation}

\paragraph{Effect of $b$ under softmax.}

Adding a logit offset $b$ to hub-related attention entries scales their unnormalized softmax weights by a factor of $e^{b}$.
In ARACH, we consider $b<0$, which down-weights hub-attention logits and reduces the fraction of attention mass assigned to these connections in the affected rows.
As a result, $b$ provides a simple calibration mechanism for hub-mediated attention, while the overall attention reallocation in ARACH is determined jointly by the hub-augmented layout and this logit scaling.

\section{Experiments}

We evaluate ARACH as a training-free inference-time plug-in on GPT-2 small using paired comparisons:
we compare the same pretrained backbone \emph{with and without ARACH applied at inference time}, while keeping model weights, decoding configuration, and the evaluation pipeline fixed.
Unless stated otherwise, both conditions use the same overall context budget; when ARACH is enabled, one position is reserved for the hub token and evaluation metrics are computed on verbal tokens only.
We perform experiments on five widely-used datasets: LAMBADA~\cite{paperno2016lambada}, PG-19~\cite{rae2019compressive}, StoryCloze~\cite{mostafazadeh2016corpus}, SQuAD~\cite{rajpurkar2016squad}, and WikiText-103~\cite{merity2016pointer}—covering cloze-style and long-form language modeling.
Additional dataset split details, task-specific evaluation protocols, and efficiency analysis are provided in the supplementary material.

\subsection{Main Results on Standard Evaluations}
\label{sec:main_results}
\paragraph{Evaluation protocol.}

We follow standard evaluation practice for each dataset, with a few task-specific details.
For LAMBADA, we use the token-level variant under GPT-2 BPE (correctness of the final target token), which is commonly reported for GPT-2 style tokenization.\footnote{The strict word-level metric can differ under subword tokenization; we adopt the token-level variant to ensure a consistent paired comparison.}
For SQuAD, we generate answers terminated by \texttt{<ANS\_END>} and report exact match and F1; to obtain meaningful answer generation, we evaluate a GPT-2 small checkpoint fine-tuned on SQuAD with and without ARACH at inference time, keeping the fine-tuned weights fixed.
For PG-19, we report sliding-window perplexity and score only the non-overlapping tail tokens in each window.We defer full dataset configurations, split construction (e.g., the held-out SQuAD split), and task-specific evaluation details to the supplementary material.

\paragraph{Main results.}

Table~\ref{tab:main_results} reports paired comparisons on the same pretrained backbone \emph{with and without ARACH applied at inference time}.
Across tasks, the consistent improvements indicate that ARACH provides a generally beneficial inference-time modification for next-token prediction, rather than a task-specific heuristic.
At the same time, the magnitude of improvement varies by evaluation: the largest gains appear on PG-19 (37.33 $\rightarrow$ 33.11 perplexity) and LAMBADA (+3.53 accuracy points), which place a stronger premium on integrating information from a broad prefix, while WikiText-103, SQuAD, and StoryCloze exhibit smaller but still positive changes.
Performance is also stable across the tested offset values, suggesting that ARACH operates in a robust regime rather than relying on brittle task-specific tuning.

Overall, these results support our hypothesis that inference-time \emph{internal} intervention can systematically improve next-token prediction, and they further suggest that the benefits are amplified when effective global context aggregation becomes more critical.
We examine this mechanism through attention-sink and routing analyses in Section~\ref{sec:sink}.

\begin{figure*}[!htb]
  \centering
  \setlength{\abovecaptionskip}{1pt}
  \setlength{\belowcaptionskip}{2pt}
  \includegraphics[width=1.00\textwidth,trim=0mm 50mm 0mm 0mm,]{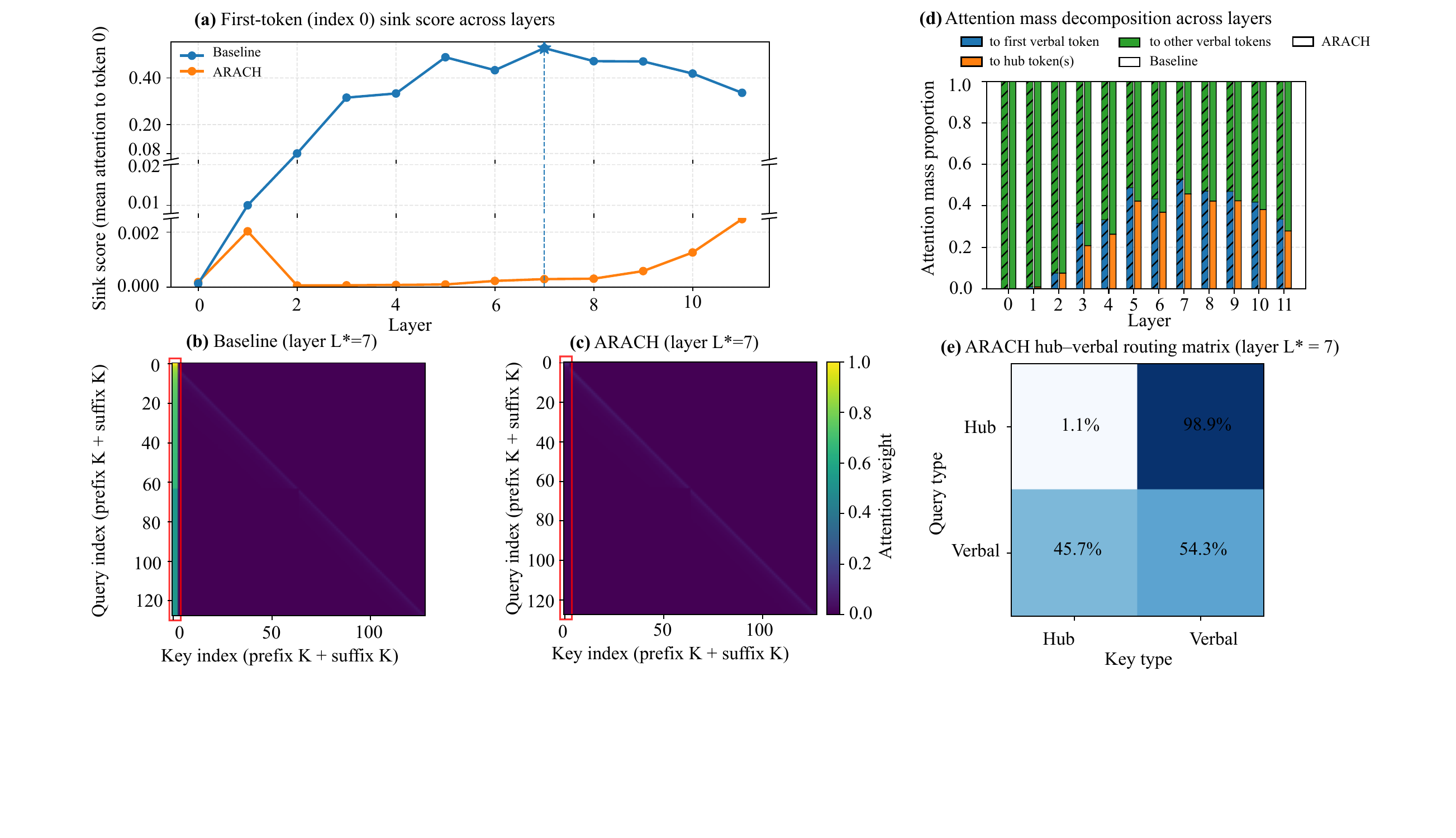}
    \caption{
    Attention sink and hub-mediated reallocation analyses on PG-19, comparing the baseline and ARACH (default logit offset setting).
    \textbf{(a)} Layerwise sink score, defined as the mean attention mass assigned to the first \emph{verbal} token, averaged over heads and samples in the test set.
    $L^*$ denotes the layer with the largest sink score under the baseline.
    \textbf{(b,c)} Heatmaps of mean attention weights among verbal tokens at layer $L^*$ for the baseline and ARACH .
    The red boxes highlight early verbal-token columns that are most associated with sink-like concentration in the baseline.
    For readability, only the first and last $K=64$ tokens are shown.
    \textbf{(d)} Layerwise attention-mass decomposition, reporting the fraction of attention assigned to the first verbal token, to hub tokens, and to the remaining verbal tokens (averaged over samples in the test set and heads).
    \textbf{(e)} Routing summary at layer $L^*$ for ARACH, reporting the fraction of attention mass allocated to each attention block. 
    }

  \label{fig:attn_sink}
\end{figure*}

\subsection{ARACH Mitigates Attention Sink}
\label{sec:sink}

\paragraph{Goal and setup.}
We analyze attention patterns to probe a hypothesis suggested by ARACH's design:
by introducing a context hub that \emph{aggregates} causally available prefix information and by \emph{regulating} hub routing via a logit offset, ARACH may reallocate attention and reduce the \textit{attention sink} phenomenon~\cite{xiao2024efficient} (i.e., over-attention to early verbal tokens).
We compute the statistics and visualizations in Figure~\ref{fig:attn_sink} on the PG-19 test set, comparing the baseline and ARACH under its default setting.

\paragraph{Locate the layer where the sink is most pronounced.}
Because sink-like concentration can vary substantially across layers, we first quantify sink strength layer-by-layer and select a representative sink-prone layer for subsequent analyses.
We compute a \emph{sink score} for each layer, defined as the mean attention mass assigned to the first verbal token.
Figure~\ref{fig:attn_sink}\textbf{(a)} shows that the baseline exhibits a clear sink peak at $L^*=7$, which we adopt as the representative layer.
In contrast, ARACH yields consistently lower sink scores across layers, which is consistent with sink mitigation.

\paragraph{Visualize verbal-token attention at the sink-prone layer.}
Having identified $L^*$, we visualize mean attention patterns among verbal tokens to examine whether the baseline exhibits the expected early-token-dominated structure and whether ARACH alters it.
Specifically, we average attention weights over samples (and heads) on the PG-19 test set and plot the resulting verbal-to-verbal attention heatmaps at layer $L^*$ for the baseline and ARACH, shown in Figure~\ref{fig:attn_sink}\textbf{(b)} and Figure~\ref{fig:attn_sink}\textbf{(c)}, respectively.
The highlighted early-token region (red boxes) emphasizes the sink-prone columns in the baseline.
Under identical aggregation and visualization settings, the baseline heatmap shows strong concentration on early verbal tokens, whereas this pattern is substantially weakened when ARACH is enabled, consistent with sink mitigation.

\paragraph{Trace where the attention mass is reallocated across layers.}
To move beyond a single-layer visualization, we decompose attention mass across layers into three parts: (i) mass assigned to the first verbal token, (ii) mass assigned to hub tokens in ARACH, and (iii) mass assigned to the remaining verbal tokens.
We compute the average proportion of attention weights assigned to these categories for both methods and summarize them as a stacked bar plot in Figure~\ref{fig:attn_sink}\textbf{(d)}.

Relative to the baseline, ARACH reduces the fraction of attention mass assigned to the first verbal token across layers.
Meanwhile, ARACH assigns a non-trivial fraction of mass to hub tokens, particularly in mid-to-late layers.
Notably, the hub-token mass under ARACH is comparable in magnitude to the baseline's first-token mass in corresponding layers, which suggests a natural interpretation: attention that would otherwise be absorbed by an early verbal position can instead be routed through the hub stream.
Since hub tokens aggregate the causally available prefix by construction, this rerouting may provide an alternative \emph{prefix-summary} conditioning path for next-token prediction, rather than relying on over-attending to a specific early verbal token.
The logit offset is designed to regulate the strength of hub routing; we study its sensitivity and role in Section~\ref{sec:bias}.

\begin{figure*}[!htb]
  \centering
  \setlength{\abovecaptionskip}{4pt}
  \setlength{\belowcaptionskip}{4pt}
  \includegraphics[width=0.95\linewidth, trim = 0mm 0mm 0mm 0mm]{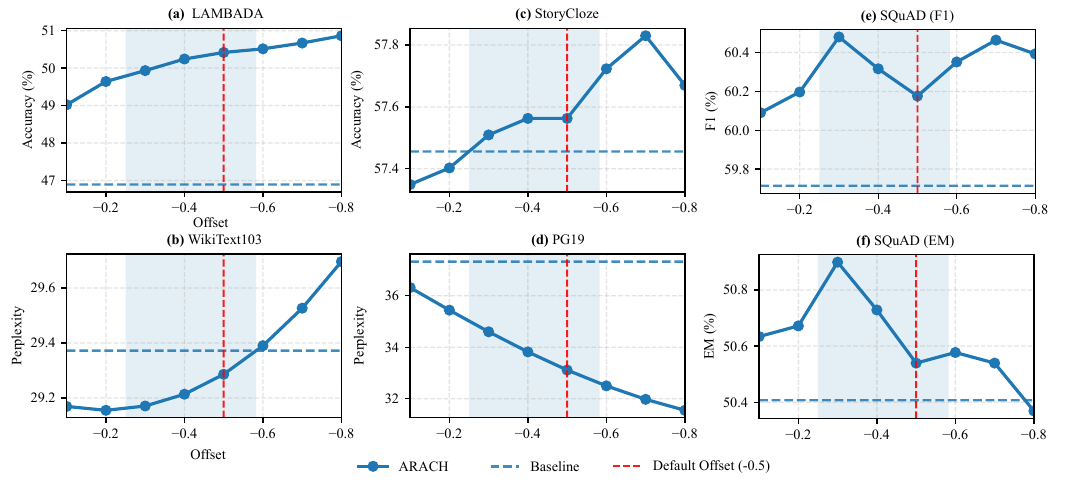}

    \caption{
    Sensitivity to the hub-attention logit offset $b$ across evaluations (sweep from $0$ to $-1.0$).
    Curves show the task metric versus $b$; the dashed horizontal line is the baseline and the red dashed vertical line marks the default $b=-0.5$.
    The blue area indicates values of $b$ for which all tasks improve over the baseline in our sweep.
    \textbf{(a)} LAMBADA (Accuracy$\uparrow$).
    \textbf{(b)} StoryCloze (Accuracy$\uparrow$).
    \textbf{(c)} SQuAD (F1$\uparrow$).
    \textbf{(d)} WikiText-103 (Perplexity$\downarrow$).
    \textbf{(e)} PG-19 (Perplexity$\downarrow$).
    \textbf{(f)} SQuAD (Exact Match$\uparrow$).
    }

  \label{fig:bias_sens}
\end{figure*}

\paragraph{Summarize routing at $L^*$ via block-wise allocation.}
Finally, to characterize routing at the representative sink-prone layer, we summarize how attention mass is distributed across the four blocks of ARACH's two-stream attention layout.
We compute the average fraction of attention weights allocated to each quadrant (hub-to-hub, hub-to-verbal, verbal-to-hub, and verbal-to-verbal) at $L^*$ and visualize the resulting block-wise allocation in Figure~\ref{fig:attn_sink}\textbf{(e)}.

Two observations follow.
First, a substantial fraction of attention mass involves the hub stream (in particular, the verbal-to-hub and hub-to-verbal blocks), indicating that hub tokens attract non-trivial attention at this sink-prone layer.
Together with the layerwise mass decomposition in Figure~\ref{fig:attn_sink}\textbf{(d)}, this is consistent with the interpretation that attention mass previously absorbed by the first verbal token in the baseline can be rerouted toward the hub pathway under ARACH, thereby alleviating sink-like concentration.
Second, the large verbal-to-hub and hub-to-verbal allocations also indicate strong cross-stream interaction, suggesting that aggregated prefix information carried by the hub can be actively exchanged with verbal tokens during prediction.
This provides a plausible mechanism for the downstream improvements: ARACH may facilitate more effective utilization of broad-prefix context by making a compact prefix summary more directly accessible through hub-mediated routing.

\begin{table*}[!t]
\centering
\setlength{\abovecaptionskip}{2pt}
\setlength{\belowcaptionskip}{2pt}
\small
\setlength{\tabcolsep}{4.0pt}
\begin{tabular}{lccccc}
\toprule
\textbf{Variant} &
\textbf{LAMBADA (Acc$\uparrow$)} &
\textbf{SQuAD (EM/F1$\uparrow$)} &
\textbf{WT-103 (PPL$\downarrow$)} &
\textbf{PG-19 (PPL$\downarrow$)} &
\textbf{StoryCloze (Acc$\uparrow$)} \\
\midrule
Baseline (no hub)     & 46.89 & 50.41/59.71 & 29.37 & 37.33 & 57.45 \\
Hub-only ($b=0$)      & 48.46 & 50.26/59.66 & \textbf{29.21} & 37.21 & 57.45 \\
Full ARACH ($b=-0.5$) & \textbf{50.42} & \textbf{50.54}/\textbf{60.18} & 29.29 & \textbf{33.11} & \textbf{57.56} \\
\midrule
$\Delta$ Hub-only     & +1.57 & -0.15/-0.05 & +0.16 & +0.12 & 0.00 \\
$\Delta$ Full         & +3.53 & +0.13/+0.47 & +0.08 & +4.21 & +0.11 \\
\bottomrule
\end{tabular}
\caption{
Ablation of the hub-attention offset term $b$ under paired evaluation.
Hub-only sets $b=0$ (hub stream retained), while Full uses the default $b=-0.5$.
We define $\Delta$ such that positive values indicate improvement.
}
\label{tab:bias_ablation}
\end{table*}

\subsection{Ablation of the Logit Offset and Sensitivity}
\label{sec:bias}



\paragraph{Component contribution and necessity.}
To isolate the role of the offset, we evaluate a \emph{hub-only} variant that keeps the hub stream but sets $b=0$.
We then compare three configurations under the same paired protocol as Table~\ref{tab:main_results}: the baseline GPT-2 small (no hub), hub-only ($b=0$), and full ARACH with the default offset ($b=-0.5$).
Table~\ref{tab:bias_ablation} shows that introducing the hub stream alone ($b=0$) can already yield improvements on some evaluations (for example, LAMBADA and WikiText-103), indicating that the hub pathway provides a useful alternative aggregation route.
However, the gains are not uniformly positive across tasks, and the improvements are modest on long-form PG-19.
In contrast, adding the offset on top of the hub stream ($b=-0.5$) yields larger and consistent gains for all tasks, with the most pronounced improvement on PG-19.
These results suggest that the hub stream is the enabling structure, while the logit offset is an important stabilizer that makes hub-mediated routing reliably beneficial; importantly, the offset is defined on hub-attention connections and therefore cannot be applied meaningfully without the hub stream.

\paragraph{Sensitivity and robust operating regime.}
Next, we sweep $b$ over a wide range and report task metrics in Figure~\ref{fig:bias_sens}.
Across tasks, ARACH exhibits a broad robust region in which performance consistently improves over the baseline (blue area), and our default choice ($b=-0.5$) lies within this region.
This stability indicates that ARACH does not require brittle task-specific tuning of $b$ to obtain gains, and is consistent with $b$ functioning as a coarse control knob for hub routing strength rather than a finely tuned hyperparameter.

\paragraph{Interpretation.}
Taken together with the attention analyses in Section~\ref{sec:sink}, the ablation results support the following interpretation: the hub stream creates a prefix-summary pathway, and the logit offset regulates how much probability mass is routed through that pathway.
A sufficiently negative offset reduces over-reliance on hub-diagonal connections, while still allowinge, enabling the hub to act as a useful intermediary for redistributing attention and making aggregated prefix information more accessible during prediction.

\section{Conclusion}
We introduced ARACH, a training-free inference-time plug-in for decoder-only Transformers that improves next-token prediction by intervening in internal attention routing.
ARACH augments each attention layer with an adaptive context hub under strict causal visibility constraints, together with a scalar hub-attention logit offset that regulates hub-mediated attention strength without updating pretrained weights.
In paired evaluations on GPT-2 small across multiple standard benchmarks, enabling ARACH consistently improves task metrics with limited additional computation.
Attention analyses further suggest that ARACH mitigates sink-like attention concentration on the first verbal token by rerouting attention through a hub-mediated pathway.
Overall, our results suggest that engineering a model's internal computation at inference time provides an orthogonal and complementary route to improving LLMs, alongside training-based post-training adaptation and prompt-based test-time scaling.

\bibliographystyle{named}
\bibliography{ijcai26}

\end{document}